\documentclass{article}

\usepackage{graphicx}      

\usepackage{bm}
\usepackage[utf8]{inputenc} 
\usepackage[T1]{fontenc}    
\usepackage{url}            
\usepackage{nicefrac}       
\usepackage{amsmath}
\usepackage{graphicx}
\usepackage{amssymb}

\newcommand{\R}{\mathbb{R}} 

\newcommand{\q}{q} 
\newcommand{\A}{A} 
\newcommand{\ac}{a} 
\newcommand{\B}{B} 
\newcommand{\bb}{b} 

\newcommand{\sens}[1]{\tilde{#1}}
\newcommand{\adjoint}[1]{\overline{#1}}

\newcommand{\tvec}[1]{\mathbf{#1}}
\newcommand{\nsamp}{T}
\newcommand{\pdiff}[2]{\frac{\partial #1}{\partial #2}}

\newcommand{\loss}{\mathcal{L}}

\newcommand{\simul}{{\rm sim}}
\newcommand{\Name}{\emph{dynoNet}}
\newcommand{\vv}{z}

\title{Deep learning with transfer functions: new applications in system identification} 
\author{%
  Dario~Piga, Marco~Forgione, Manas~Mejari \\
  IDSIA Dalle Molle Institute for Artificial Intelligence\\
  USI-SUPSI, Lugano, Switzerland\\
  \texttt{\{dario.piga, marco.forgione, manas.mejari\}@supsi.ch} \\
}

\begin{document}

\maketitle

\begin{abstract}                
This paper presents a linear dynamical operator described in terms of a rational transfer function, endowed with a well-defined and efficient back-propagation behavior for automatic derivatives computation.
The operator enables end-to-end training of structured networks containing linear transfer functions and other differentiable units {by} exploiting standard deep learning software.
 Two relevant applications of the operator in system identification are presented. The first one consists in the  integration of \emph{prediction error methods} in deep learning. The dynamical operator is included as {the} last layer of a neural network in order to obtain the optimal one-step-ahead prediction error. 
 The second one considers identification of general block-oriented models from quantized data. These block-oriented models are constructed by combining linear dynamical operators with static nonlinearities described as standard feed-forward neural networks. A custom loss function corresponding to the log-likelihood of quantized output observations is defined. For gradient-based optimization, the derivatives of the log-likelihood are computed by applying the back-propagation algorithm through the whole network.  Two system identification benchmarks are used to show the effectiveness of the proposed methodologies. 
\end{abstract}

\vskip 1em 
\noindent\rule{\textwidth}{1pt}
 To cite this work, please use the following bibtex entry:
\begin{verbatim}
@inproceedings{piga2021a,
  title={Deep learning with transfer functions:
            new applications in system identification},
  author={Piga, D. and Forgione, M. and Mejari, M.},
  booktitle={Proc. of the 19th IFAC Symposium System Identification:
                learning models for decision and control},
  address={Padova, Italy},
  year={2021}
}
\end{verbatim}
\vskip 1em
Using the plain bibtex style, the bibliographic entry should look like:\\ \\
\textsf{D. Piga, M. Forgione and M. Mejari Deep learning with transfer functions: new applications in system identification.} 
 In \textit{Proc. of the 19th IFAC Symposium System Identification: learning models for decision and control}, 
 Padova, Italy, 2021.

\noindent\rule{\textwidth}{1pt}



\section{Introduction}


Thanks to  the universal approximation capabilities of neural networks, deep learning algorithms are nowadays behind high-impact cutting-edge technologies such as language translation, speech recognition, and autonomous driving, just to cite a few \cite{schmidhuber2015deep}. Highly optimized and user-friendly deep learning frameworks are available \cite{paszke:2017automatic}, often distributed under permissive open-source licenses.
On one hand, using the high-level functionalities offered by a deep learning framework, standard learning tasks (once considered extremely hard) such as image classification can be accomplished with moderate efforts even by non-expert practitioners. On the other hand, experienced users can build customized models and objective functions,
exploiting the framework's built-in back-propagation engine \cite{rumelhart1988learning} for gradient computations. 

\emph{System identification} definitely represents a challenging field where the flexibility of neural networks and  deep learning algorithms can be used to describe and estimate non-linear dynamical systems. This requires {one} to define specialized networks that take into account  temporal evolution of the input/output data, a fundamental feature setting identification of dynamical systems apart from typical (static) supervised problems in machine learning.   

Neural networks have been widely used in system identification in the past (see, e.g., 
 \cite{werbos1989neural, chen1990non,chen1992neural}), and the \emph{back-propagation through time} algorithm \cite{williams1995gradient} has been applied to train \emph{Recurrent Neural Networks} (RNNs). Recently, the use of 
 \emph{Long Short-Term Memory} (LSTM)  and 1-D  \emph{Convolutional Neural Networks} (CNNs) in system identification has been discussed in~\cite{gonzalez2018non, wang2017new} and~\cite{andersson:2019deep}, respectively.  
 An architecture specialized for continuous-time system identification called \emph{Integrated Neural Network}, 
 {which consists of} a feed-forward network followed by an integral block is proposed in~\cite{Mav2020}.  Loss functions based on the simulation error over small subsets of the training data are proposed in~\cite{forgione2019model,forgione2021continuous} and~\cite{ribeiro_smoothness_2020}, where a regularization  term and a multiple shooting approach are employed, respectively, to enforce 
   that the initial conditions of all the subsets are compatible with the identified model dynamics. 
   The list above is far from being exhaustive, as new contributions using neural networks and deep learning algorithms for system identification are regularly appearing   every year in dedicated conferences and journals.

  A novel network architecture called \Name \ {which is}  tailored for the identification of nonlinear dynamical systems has been recently proposed by the authors in~\cite{forgione2021dynonet}. The network consists in the interconnection of \emph{Linear Time-Invariant} (LTI) dynamical blocks and static nonlinearities.   
   In this paper, we describe the LTI dynamical layer which constitutes the elementary block of \Name, along with the  forward and backward operations needed to make this layer compatible with the back-propagation algorithm. The LTI layer is described  in terms of rational transfer functions, and thus acts as an \emph{infinite impulse response} (IIR) filter applied to its input sequence. 

    A differentiable LTI dynamical layer allows us to tackle several challenging problems in system identification. In particular,  in this paper:
  \begin{itemize}
  	\item   We consider the case of output signals affected by an additive colored noise. To this aim, we include  a trainable linear dynamical unit as {the} last layer of an end-to-end differentiable network representing a dynamical system (e.g., a convolutional, recurrent, or \Name\ network) in order to build the ``one-step-ahead prediction error'' minimized in the popular \emph{Prediction Error Method} (PEM).  Such an application of PEM in deep learning will be referred to as \emph{deepPEM};
  	\item We address the problem of identification from quantized output observations, and present a method 
  	that is applicable to end-to-end differentiable model structures (including \Name), by properly changing the likelihood function maximized in training. As  \Name \ represents a 
  	generalization of classic \emph{block-oriented} architectures~\cite{giri2010block}, the proposed approach can be applied for {the} identification of block-oriented models from quantized data.
  \end{itemize}

  The linear dynamical operator has been implemented in the  \emph{PyTorch} deep learning framework
   and the software is available for download at  \url{https://github.com/forgi86/sysid-transfer-functions-pytorch}.

The rest of this paper is organized as follows. The linear dynamical operator is described in Section \ref{sec:LTI} and the steps required to integrate it in a deep learning framework are described in Section \ref{sec:layer_DL}. The \emph{deepPEM} algorithm and the problem of identification from quantized data are discussed in Section \ref{sec:pem} and \ref{sec:quantized}, respectively. Finally, two benchmark  examples are presented in Section \ref{sec:examples}.

\section{Dynamical layer} \label{sec:LTI}
The input-output relation of the linear dynamical layer is described by a rational transfer function $G(q)$ as:
	\begin{equation} \label{eqn:filter} 
	y(t) = G(\q) u(t) =  \frac{\B(\q)}{\A(\q)}u(t), 
	\end{equation}
	where 	  $u(t) \in \R$ and  $y(t) \in \R$ are the input and output signals of the filter $G(q)$ at time  $t$ {respectively}, 	and 
	 $A(q)$ and $B(q)$ are polynomials in the \emph{time delay operator}  $\q^{-1}$ ($q^{-1}u(t)=u(t-1)$), i.e., 
\begin{subequations}
	\begin{align}
	\A(q) &= 1 + \ac_1 \q^{-1} + \dots + \ac_{n_\ac}q^{-n_\ac}, \\
	\B(q) &= \bb_0 + \bb_1 \q^{-1} + \dots + \bb_{n_{\bb}}q^{-n_{\bb}}.
	\end{align}
\end{subequations}

The coefficients   of the polynomials $\A(q)$ and $\B(q)$ are collected in vectors  $\ac = [\ac_1\; \ac_2\dots\;\ac_{n_\ac}] \in \R^{n_\ac}$ and $\bb = [\bb_0\; \bb_1\; \dots \;\bb_{n_\bb}] \in \R^{n_\bb + 1}$. They represent 
 the tunable parameters of the filter $G(q)$.

The filtering operation  in \eqref{eqn:filter} has the following  input/output representation:
\begin{equation} \label{eqn:filterA}
\A(\q)y(t) =  \B(\q)u(t),
\end{equation}
which, according to the definitions of $\A(q)$ and $\B(q)$, is equivalent to the linear difference  equation:
\begin{align}
\label{eq:OE_predictor_b}
y(t) = & -\ac_1 y(t\!-\!1) \dots - \ac_{n_\ac} y(t\!-\!n_\ac) + \nonumber \\
& + \bb_0 u(t) + \bb_1 u(t-1) + \dots + \bb_{n_\bb}\!u(t-n_\bb).
\end{align}

 In the following, we assume that the filter $G(\q)$ is always initialized from rest, i.e., $u(t)=0$ and $y(t)=0$ for $t < 0$.

 Let us stack the input and output samples $u(t)$ and $y(t)$ from time $0$ to $\nsamp\!-\!1$ in vectors $\tvec{u} \in \mathbb{R}^{\nsamp}$  and $\tvec{y} \in \mathbb{R}^{\nsamp}$, respectively.\footnote{In the following, the bold-face notation is reserved to real-valued $\nsamp$-length vectors.  For instance, $\tvec{u} \in \R^{\nsamp}$ is a $T$-length vector with entries $\tvec{u}_0, \tvec{u}_1, \dots, \tvec{u}_{\nsamp -1}$.} 
 With a slight abuse of notation, the filtering operation in \eqref{eqn:filter} applied to $\tvec{u}$ is denoted as 
 $
 \tvec{y} = G(\q)\tvec{u}$. 
 This operation is also equivalent to the convolution 
 \begin{equation}
 \label{eq:G_conv}
 \tvec{y}_i =  \sum_{j=\max{(0, i+1-T)}}^{\min(i, T-1)} \tvec{g}_j \tvec{u}_{i-j},\qquad  
 i=0,1,\dots,\nsamp\!-\!1,
 \end{equation}
 where $\tvec{g} \in \R^{\nsamp}$ is a vector containing the first $\nsamp$ samples of the  \emph{impulse response} of $G(q)$.

The derivations are presented in the paper for a proper, single-input-single-output (SISO) transfer function to simplify the notation. Note that the software implementation available in the paper's on-line repository 
allows setting an arbitraty number of input delays $n_k$, {\emph{i.e.} $\B(q) = \bb_0 \q^{-n_k} + \bb_1 \q^{-n_k-1} +  \dots + \bb_{n_{\bb}}q^{-(n_k-n_{\bb})}$ with $n_k \geq 0$}, and includes support for the multi-input-multi-output (MIMO) case.

\section{Dynamical Layer in Deep Learning}
\label{sec:layer_DL}

In this section, the forward and backward pass operations required to integrate the linear dynamical layer in a deep learning framework are derived. The operator $G(q)$ interpreted as a differentiable block for use in deep learning will be referred to as $G$-block in the rest of the paper.

\subsection{Forward pass}
\begin{figure}
	\begin{center}
		\includegraphics[width=140pt]{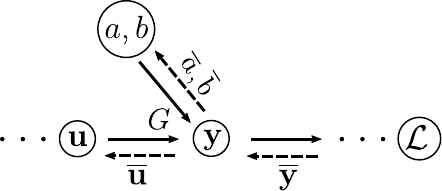}
	\end{center}
	\caption{Forward and backward operations of a $G$-block within a computational graph.}
	\label{fig:backprop_tf}
\end{figure}

The forward operations of a $G$-block are represented by solid arrows  in the computational graph sketched in Fig.~\ref{fig:backprop_tf}. 
In the {forward pass}, the block filters an input sequence $\tvec{u} \in \mathbb{R}^{\nsamp}$ through a dynamical system $G(\q)$ with  parameters $\ac = [\ac_1\;\dots\;\ac_{n_\ac}]$ and $\bb = [\bb_0\; \bb_1\; \dots \;\bb_{n_\bb}]$. The block output   $\tvec{y} \in \mathbb{R}^{\nsamp}$ contains the filtered sequence:
\begin{equation}
\label{eq:forward_op}
\tvec{y} =  G(\q,b,a) \tvec{u}.
\end{equation}
The input $\tvec{u}$ of the $G$-block  may be either the training input sequence or the result of previous operations in the computational graph, while
the output $\tvec{y}$ is an intermediate step towards the computation of a scalar loss $\loss$. 

When the filtering operation \eqref{eq:forward_op} is implemented through the recurrent equation \eqref{eq:OE_predictor_b}, the computational cost of the forward pass for the  $G$-block corresponds to $\nsamp(n_\bb + n_\ac + 1)$ multiplications. These multiplications  need to be performed sequentially for the $\nsamp$ time samples, but can be parallelized for the $n_\bb$ + $n_\ac + 1$ different coefficients at each time index $t$.

\subsection{Backward pass}
The backward operations are illustrated in Fig.  \ref{fig:backprop_tf} with dashed arrows. 
In the {backward pass}, the $G$-block receives the vector $\adjoint{\tvec{y}} \in \mathbb{R}^{\nsamp}$ with the partial derivatives of the loss $\loss$ w.r.t. $\tvec{y}$, i.e., 
\begin{equation}
\adjoint{\tvec{y}}_t = \pdiff{\loss}{\tvec{y}_t},\quad  t=0,\dots,\nsamp-1,  
\end{equation}
and  it has to compute the derivatives of the loss $\loss$ w.r.t. its differentiable inputs  $\bb$, $\ac$, and $\tvec{u}$, i.e., 

\begin{subequations}
	\begin{align} 
	\adjoint{\bb}_j &= \pdiff{\loss}{\bb_j},\qquad j=0,\dots,n_\bb \\
	\adjoint{\ac}_j &= \pdiff{\loss}{\ac_j},\qquad j=1,\dots,n_\ac\\
	\adjoint{\tvec{u}}_\tau &= \pdiff{\loss}{\tvec{u}_\tau},\qquad \tau=0,1,\dots,\nsamp\!-\!1. \label{eq:adjsensu}
	\end{align}
\end{subequations}

\subsubsection{{Derivatives w.r.t} numerator coefficients $\bb$.} 

Application of the chain rule leads to:
\begin{equation}
\label{eq:backprop_b_first}
\adjoint{\bb}_j = \sum_{t=0}^{\nsamp-1} \pdiff{\loss}{\tvec{y}_t} \pdiff{\tvec{y}_t}{\bb_j} = \sum_{t=0}^{\nsamp-1} \adjoint{\tvec{y}}_t \pdiff{\tvec{y}_t}{b_j}
\end{equation}

The  sensitivities  $\sens{\bb}_j(t)  = \pdiff{\tvec{y}_t}{b_j}$, $j=0,1,\dots,n_\bb,$ can be computed through recursive filtering operations~\cite{ljung:1999system}. Specifically,  by differentiating 
the left  hand side of Eq. \eqref{eqn:filterA} w.r.t $\bb_j$ we obtain:
\begin{align}  \label{eq:sens_b}
\A(\q) \sens{\bb}_j(t)  = u(t-j),
\end{align}

Thus, $\sens{\bb}_j(t)$ can be computed by filtering the input vector $u(t)$  through the linear filter $ \displaystyle \frac{1}{\A(\q)}$. Furthermore, 
\begin{equation}
\label{eq:regime_b2}
\sens{\bb}_{j}(t) = \begin{cases}
\sens{\bb}_0(t-j), \;\;&t-j \geq 0\\
0,           \;\; & t-j < 0.
\end{cases}
\end{equation}
Then,  only $ \sens{\bb}_0(t)$ needs to be recursively simulated through Eq.~\eqref{eq:sens_b}. The  sensitivities $ \sens{\bb}_j(t)$, $j=1,\ldots,n_\bb$, are  computed  through simple shifting operations. 

From~\eqref{eq:backprop_b_first} and~\eqref{eq:regime_b2}, the $j$-th component of $\adjoint{\bb}$ is given by:
\begin{equation}
\label{eq:backprop_b_last}
\adjoint{\bb}_j = \sum_{t=j}^{\nsamp-1} \adjoint{\tvec{y}}_t \sens{b}_0(t-j).
\end{equation}

Overall, the  computation of  $\adjoint{\bb}$ requires to: 
\begin{itemize}
	\item filter the input $\tvec{u}$ through $\frac{1}{A(\q)}$, which entails  $\nsamp n_\ac$ multiplications.  Because of the recursive form ~of \eqref{eq:sens_b}, these operations need  to be performed sequentially;
	\item compute 
	the $n_{\bb}\!+\!1$ dot products  in \eqref{eq:backprop_b_last}, for a total of  $\nsamp(n_{\bb}\!+\!1)- n_\bb$ multiplications. These operations can be parallelized.
\end{itemize}

\subsubsection{{Derivatives w.r.t} denominator coefficients $\ac$.} 
The   sensitivities $\sens{\ac}_j(t) = \pdiff{\tvec{y}_t}{\ac_j}$, $j=1,2,\dots,n_\ac$ are obtained based on the same rationale described above, 
by differentiating the terms in  Eq. \eqref{eqn:filterA} with respect to $\ac_j$. This yields:
\begin{equation}
\label{eq:sens_a}
\sens{\ac}_j(t) = -\frac{1}{\A(\q)}y(t-j).
\end{equation}

The following condition holds:
\begin{equation}
\label{eq:regime_a2}
\sens{\ac}_{j}(t) = \begin{cases}
\sens{\ac}_1(t-j+1), \;\;&t-j+1 \geq 0\\
0,           \;\; & t-j+1 < 0
\end{cases}
\end{equation}
and the $j$-th component of $\adjoint{\ac}$ is given by: 
\begin{equation} \label{eqn:abar}
\adjoint{\ac}_j = \sum_{t=j-1}^{\nsamp-1} \adjoint{\tvec{y}}_t \sens{a}_1(t-j+1).
\end{equation}

The back-propagation for the  coefficients $a$ thus requires: ($i$) the  filtering operation \eqref{eq:sens_a}, which involves  $\nsamp n_\ac$ multiplications; and ($ii$) 
the $n_\ac$ dot products defined in \eqref{eqn:abar}, for a total of $\nsamp n_\ac - n_\ac + 1$ multiplications. 

\subsubsection{Derivatives w.r.t. input time series  $\tvec{u}$.} 
In order to compute the partial derivatives $\pdiff{\loss}{\tvec{u}_\tau} $ the  chain rule is applied, which yields:
\begin{equation}
\label{eq:chainrule_u}
\adjoint{\tvec{u}}_\tau = \pdiff{\loss}{\tvec{u}_\tau} 
= \sum_{t=0}^{\nsamp-1}{\pdiff{\loss}{\tvec{y}_t} \pdiff{\tvec{y}_t}{\tvec{u}_\tau}} 
= \sum_{t=0}^{\nsamp-1}{\adjoint{\tvec{y}}_t \pdiff{\tvec{y}_t}{\tvec{u}_\tau}} 
\end{equation}
From \eqref{eq:G_conv}, the term $\pdiff{\tvec{y}_t}{\tvec{u}_\tau}$ is given by:
\begin{equation} \label{eqn:dydu}
\pdiff{\tvec{y}_t}{\tvec{u}_\tau} = \begin{cases}
\tvec{g}_{t-\tau},\;\; &t-\tau \geq 0\\
0, \; & t-\tau < 0.
\end{cases}
\end{equation}
By substituting Eq.~\eqref{eqn:dydu} into \eqref{eq:chainrule_u}, we obtain 
\begin{equation} \label{eq:backprop_u_corr2}
\adjoint{\tvec{u}}_\tau = 
\sum_{t=\tau}^{\nsamp-1}\adjoint{\tvec{y}}_t 
\tvec{g}_{t-\tau} 
\end{equation}

Direct implementation of the cross-correlation~\eqref{eq:backprop_u_corr2} requires a number of operations which grows \emph{quadratically} with   $\nsamp$. 
A more efficient solution is obtained by observing:
\begin{align*}
& \adjoint{\tvec{u}}_0 = \sum_{t=0}^{\nsamp-1} \adjoint{\tvec{y}}_t \tvec{g}_t, \ \  \ \ 
\adjoint{\tvec{u}}_1 = \sum_{t=1}^{\nsamp-1} \adjoint{\tvec{y}}_t \tvec{g}_{t-1}, \ \ \ \ 
\adjoint{\tvec{u}}_2 = \sum_{t=2}^{\nsamp-1} \adjoint{\tvec{y}}_t \tvec{g}_{t-2}, \\
&\dots, \ \ \ \  \adjoint{\tvec{u}}_{\nsamp-2} = \adjoint{\tvec{y}}_{\nsamp-2} \tvec{g}_0 + \adjoint{\tvec{y}}_{\nsamp-1}\tvec{g}_1, \ \ \ \ \ 
\adjoint{\tvec{u}}_{\nsamp-1} = \adjoint{\tvec{y}}_{\nsamp-1} \tvec{g}_{0}.
\end{align*}

Since $\tvec{g}$ represents the impulse response of  $G(q)$, the vector $\adjoint{\tvec{u}}$ can be obtained  by filtering the vector $\adjoint{\tvec{y}}$ in reverse time through $G(q)$, and then reversing the result, i.e., 
\begin{equation}
\label{eq:backprop_u_filt}
{\adjoint{\tvec{u}}} = \textrm{flip}\big(G(q)\textrm{flip}(\adjoint{\tvec{y}})\big),
\end{equation}
where $\textrm{flip}(\cdot)$ denotes the time reversal operator applied to a $\nsamp$-length vector, defined as
\begin{equation}
\left(\textrm{flip}(\tvec{x})\right)_t = \tvec{x}_{\nsamp-t-1},\qquad t=0,1,\dots,\nsamp-1. 
\end{equation}

Eq.~\eqref{eq:backprop_u_filt} represents the filtering of a $\nsamp$-length vector through $G(\q)$, whose  complexity grows  linearly with  $\nsamp$.

\section{Applications in system identification}

\subsection{deepPEM} 
\label{sec:pem}
As a first application of the differentiable LTI layer discussed in the previous section, we  integrate PEM in a deep learning context. To this aim, let us consider the data-generating system   which provides the output $y$ at time $t$ according to the equation:
\begin{align}
\label{eq:truesys_PEM}
y(t) = \mathcal{S}(U_t) + \overbrace{H_{\rm o}(q) e(t)}^{=v(t)},
\end{align}
where $\mathcal{S}(\cdot)$ is the deterministic causal component processing the entire past input sequence $U_t$ up to time $t$, and $v(t)$ is an additive
colored noise source obtained by filtering a white noise $e(t)$ through a stable linear filter 
$H_{\rm o}$. According to the PEM framework~\cite[Ch. 2.3]{ljung:1999system}, we assume that $H_{\rm o}$ is monic  and minimum phase.



Let us now consider a model structure
\begin{align}
\label{eq:model_PEM}
y(t) = \mathcal{M}(U_t, \theta_\mathcal{M}) + H(q, \theta_H) e(t),
\end{align}
where $\mathcal{M}(U_t,\theta_{\mathcal{M}})$ is a deterministic causal component (described by the parameter vector $\theta_{\mathcal{M}}$) modeling the deterministic component  $\mathcal{S}(U_t)$ in~\eqref{eq:truesys_PEM}  
and $H(q, \theta_H)$ is a transfer function with parameters $\theta_H = [a_H \ b_H]$. We denote with $\theta = [\theta_{\mathcal{M}} \ \theta_{H}]$ the vector containing all the unknown parameters in~\eqref{eq:model_PEM}. 

As known (see ~\cite[Ch. 2.3]{ljung:1999system}), the optimal \emph{one-step-ahead predictor} for  \eqref{eq:model_PEM} at time $t$, given input and output data up to time $t-1$, is given by
\begin{multline} \label{eqn:y_pred}
\hat y(t|t-1) = H^{-1}(q, \theta_H) \mathcal{M}(U_t, \theta_\mathcal{M}) \\+ [1\!-\!H^{-1}(q, \theta_H)]y(t).
\end{multline}
{Note that the output predictor in~\eqref{eqn:y_pred}  only depends on outputs up to time $t-1$ since $\!H^{-1}(q, \theta_H)$ is monic.}
The  \emph{prediction error}  $\varepsilon(t) = y(t) - \hat y(t|t-1)$ is then given by
\begin{equation}
\varepsilon(t, \theta) = H^{-1}(q, \theta_H)(y(t) - \mathcal{M}(U_t, \theta_\mathcal{M})).
\end{equation}
According to the Prediction Error Minimization (PEM) criterion, we consider the loss function:
\begin{equation}
\label{eq:loss_PEM}
\mathcal{L}(\theta) = \frac{1}{\nsamp} \sum_{t=0}^{\nsamp -1} \ell(\bm{\varepsilon}_t(\theta)),
\end{equation}
where $\bm{\varepsilon} \in \mathbb{R}^\nsamp$ is the  vector stacking the prediction error $\varepsilon$ from time $0$ up to  $\nsamp - 1$, and 
 $\ell(\cdot)$ is a scalar-valued function (e.g., $\ell(\bm{\varepsilon}_t(\theta))=\bm{\varepsilon}^2_t(\theta)$).

The differentiable transfer function discussed in this paper enables easy implementation of the PEM 
framework for models where the deterministic component $\mathcal{M}(\cdot,\theta_\mathcal{M})$ is   compatible with back-propagation. For instance, $\mathcal{M}(\cdot,\theta_\mathcal{M})$ could be a RNN, a 1-D CNN, or a \Name. 

\begin{figure}
	\centering
	\includegraphics[width=.45\textwidth]{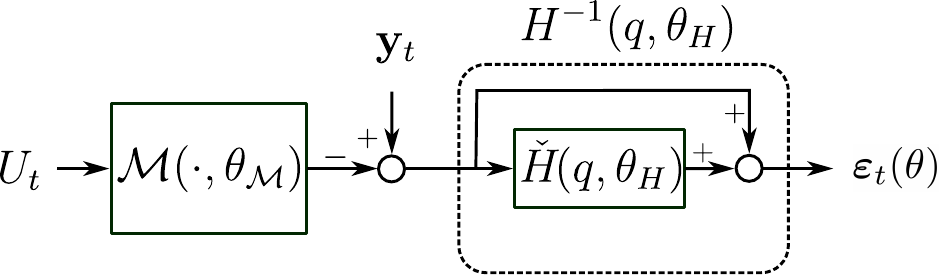} \vspace{-0.1cm}
	\caption{\emph{deepPEM}: block diagram representing the computation of the one-step-ahead prediction error.}
	\label{fig:neural_PEM}
\end{figure}

The corresponding computational graph for \emph{deepPEM} is outlined in Fig. \ref{fig:neural_PEM}. Note that the monic transfer function $H^{-1}(q, \theta)$ is constructed as $$H^{-1}(q, \theta_H) = 1 + \check{H}(q, \theta_H),$$
where $\check{H}(q, \theta_H)$ is a strictly proper transfer function implemented as 
a $G$-block with one input delay ($n_k\!=\!1$). 
This makes the estimated $H^{-1}(q, \theta_H)$ monic by construction.

\subsection{Learning from quantized observations} 
\label{sec:quantized}
As a second application, we consider the problem of learning dynamical systems from \emph{quantized} output measurements. Specifically, we assume that at each time $t$ we observe a quantized output $\tvec{\vv}_t \in \{0,1,\ldots,K\!-\!1\}$. The integer $K$ represents the number of quantization intervals and $\tvec{\vv}_t$ is given by
\begin{subequations}
\begin{align}
\label{datagen_quant}
 \tvec{\vv}_t &= Q(\tvec{y}_t + \tvec{e}_t),
\end{align}
\end{subequations}
where $\tvec{y}_t$ is the latent non-quantized output of the underlying data generating system, $\tvec{e}_t$ is a zero-mean white Gaussian noise with (unknown) standard deviation $\sigma_e$, and 
 $Q(\cdot)$ is the quantizing operator, defined as
\begin{equation}
 Q(x) = m \;\; \text{if}\;\; x \in (q_m, q_{m+1}],
\end{equation}
with  $(q_m, q_{m+1}]$ denoting  the $m$-th quantization interval.

Let us consider a simulation model $\mathcal{M}(\cdot, \theta_\mathcal{M})$ parametrized by a vector $\theta_\mathcal{M}$ which provides an output   $\tvec{y}^{\simul}_t$  when fed with an input sequence $U_t$, i.e.,  $\tvec{y}^{\simul}_t = \mathcal{M}(U_t, \theta_\mathcal{M})$. 

Because of the conditional independence of the observed outputs $\tvec{\vv}_t$ given the non-quantized output
$\tvec{y}_t$, the log-likelihood function of the parameters $\theta = [\theta_\mathcal{M}\ \sigma_e]$ is: 
\begin{align} 
\label{eq:likelihood_quant}
\mathcal{L}(\theta) =  &  \sum_{t=0}^{T-1} \log \left[  p( Q(\tvec{y}^{\simul}_t + \tvec{e}_t) = \tvec{\vv}_t  )  \right] \nonumber  \\
 =  &  \sum_{t=0}^{T-1} \log \left[ p( q_{\tvec{\vv}_t} <  \tvec{y}^{\simul}_t +  \tvec{e}_t \leq q_{\tvec{\vv}_{t+1}} )  \right] \nonumber \\
=  & \sum_{t=0}^{T-1} \underset{\mathcal{L}_t(\theta)} {\underbrace{\log \left[ \Phi\left(\frac{ q_{\tvec{\vv}_{t+1}}-\tvec{y}^{\simul}_t}{\sigma_e} \right)
\!-\! \Phi\left(\frac{ q_{\tvec{\vv}_t}\!-\!\tvec{y}^{\simul}_t}{\sigma_e} \right) \right]}},  
\end{align} 
where $\Phi$ is the cumulative density function of the Normal distribution, i.e., 
$\Phi(x) = \int_{w=-\infty}^x \mathcal{N}(w; 0, 1) dw$.  

%

\begin{figure}
 \centering
 \includegraphics[width=.4\textwidth]{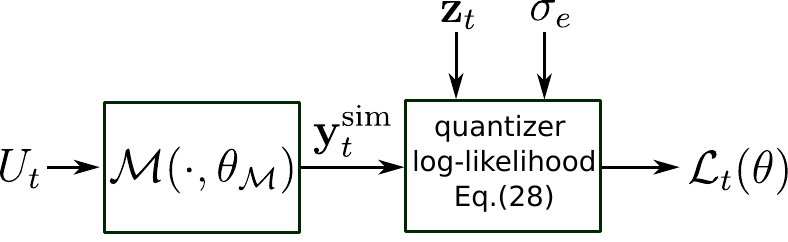}\vspace{-0.0cm}
 \caption{Learning from quantized data: block diagram representing the computation of the log-likelihood $\mathcal{L}_t(\theta)$~\eqref{eq:likelihood_quant}  of the quantized output sample $\tvec{\vv}_t$.}
 \label{fig:quantized}
\end{figure}

The log-likelihood \eqref{eq:likelihood_quant} may be readily implemented in 
current deep learning software frameworks using standard differentiable blocks. The block diagram corresponding to the log-likelihood $\mathcal{L}_t(\theta)$ of a single quantized observation $\tvec{\vv}_t$ is sketched in  Fig.~\ref{fig:quantized}. The model   $\mathcal{M}(\cdot, \theta_{\mathcal{M}})$ can be a \Name\ network, which allows us to train   block-oriented models  (constituted by trainable $G$-blocks followed by static nonlinearities)
from quantized output  observations. 

%

\section{Examples}
\label{sec:examples}
The effectiveness of the proposed methodologies is evaluated on benchmark datasets for system identification  available at the website \url{www.nonlinearbenchmark.org}. We present, in particular, results based on the Wiener-Hammerstein (WH) and the parallel Wiener-Hammerstein (PWH)   benchmarks described in  \cite{ljung2009wiener} and \cite{schoukens2015parametric}, respectively. 
The original training datasets of the two benchmarks are modified to make the identification problem  more challenging and to highlight flexibility of the presented methodologies   in handling non-standard learning problems. For the WH benchmark, we perturb the output with an additive colored noise and adopt the \emph{deepPEM} method described in Section \ref{sec:pem} to   estimate the WH model parameters and the power spectrum of the noise.  For the PWH benchmark,  we consider the case of quantized output measurements and estimate the model parameter by maximizing the log-likelihood \eqref{eq:likelihood_quant}. 

We use \Name\ networks to describe the deterministic system component in the two benchmarks, with 
an obvious choice of the network architectures reflecting the block-oriented structures of the  PWH and WH systems.

All computations are carried out on a PC equipped with an AMD Ryzen 5 1600x processor and 32 GB of RAM.
 The codes required to reproduce the results   are available in the GitHub repository \url{https://github.com/forgi86/sysid-transfer-functions-pytorch.git}.  The   reader is referred to~\cite{forgione2021dynonet} for further examples showing the effectiveness of \Name.

\subsection{Training settings and metrics}
The Adam algorithm \cite{kingma2014adam} is used for gradient-based optimization. The number $n$ of iterations  is chosen sufficiently large to reach a cost function plateau. The learning rate $\lambda$ is adjusted by a rough trial and error. All static non-linearities   are modeled as feed-forward Neural Networks with a single hidden layer containing 10 neurons and hyperbolic tangent activation function. 


The $\mathrm{fit}$ index and the Root Mean Square Error (RMSE) are used to assess the quality of the identified models: 
\begin{footnotesize}
	\begin{align*}
 \mathrm{fit} &= 100\cdot \left(1- \frac{\sqrt{\sum_{t=0}^{\nsamp-1} \left(\tvec{y}_t - \tvec{y}^{\simul}_t\right)^2} }  
{\sqrt{\sum_{t=0}^{\nsamp-1} \left(\tvec{y}_t -  {\overline{{\tvec{y}}}}\right)^2}}\right) (\%), 
\\ 
\mathrm{RMSE} &= \sqrt{\frac{1}{\nsamp} \sum_{t=0}^{\nsamp-1} \left(\tvec{y}_t - \tvec{y}^{\simul}_t\right)^2}, 
\end{align*}
\end{footnotesize}
where $\tvec{y}^{\simul}_t$ is the open-loop simulated output of the estimated  model   
and $\overline{\tvec{y}}$ is the average   of the measured output sequence.  Both the $\mathrm{fit}$ and the RMSE indexes   are measured  on the benchmarks' original test data.

\subsection{WH with colored noise}

The experimental setup used in this benchmark is an electronic circuit described in \cite{ljung2009wiener} that behaves as a Wiener-Hammerstein system. Therefore, a simple \Name\ architecture corresponding to the WH model structure is adopted. Specifically, the \Name\ model used has a sequential structure defined by a SISO $G$-block with $n_a = n_b = 8, n_k=1$; a SISO feed-forward neural network; and a final SISO $G$-block with $n_a = n_b = 8, n_k=0$. 

The original training dataset is modified by adding to the measured output 
a colored noise $v$ with standard deviation $0.1$ obtained by filtering a white noise $e$ through the transfer function 
$H_{\rm o}(\q) = \frac{1 - 1.568\q^{-1} + 0.902\q^{-2}}{1 - 1.901 \q^{-1} + 0.9409\q^{-2}}$.

In order to jointly estimate the system and the noise disturbance, the \emph{deepPEM} approach presented in Section 
\ref{sec:pem} is applied.
 The model is trained over $n=40000$ iterations of the Adam algorithm with learning rate $\lambda=10^{-4}$ on the whole training dataset ($\nsamp=100000$ samples). The total training time is 267 seconds. 
  
\begin{figure}
	\centering
	\includegraphics[width=.7\textwidth]{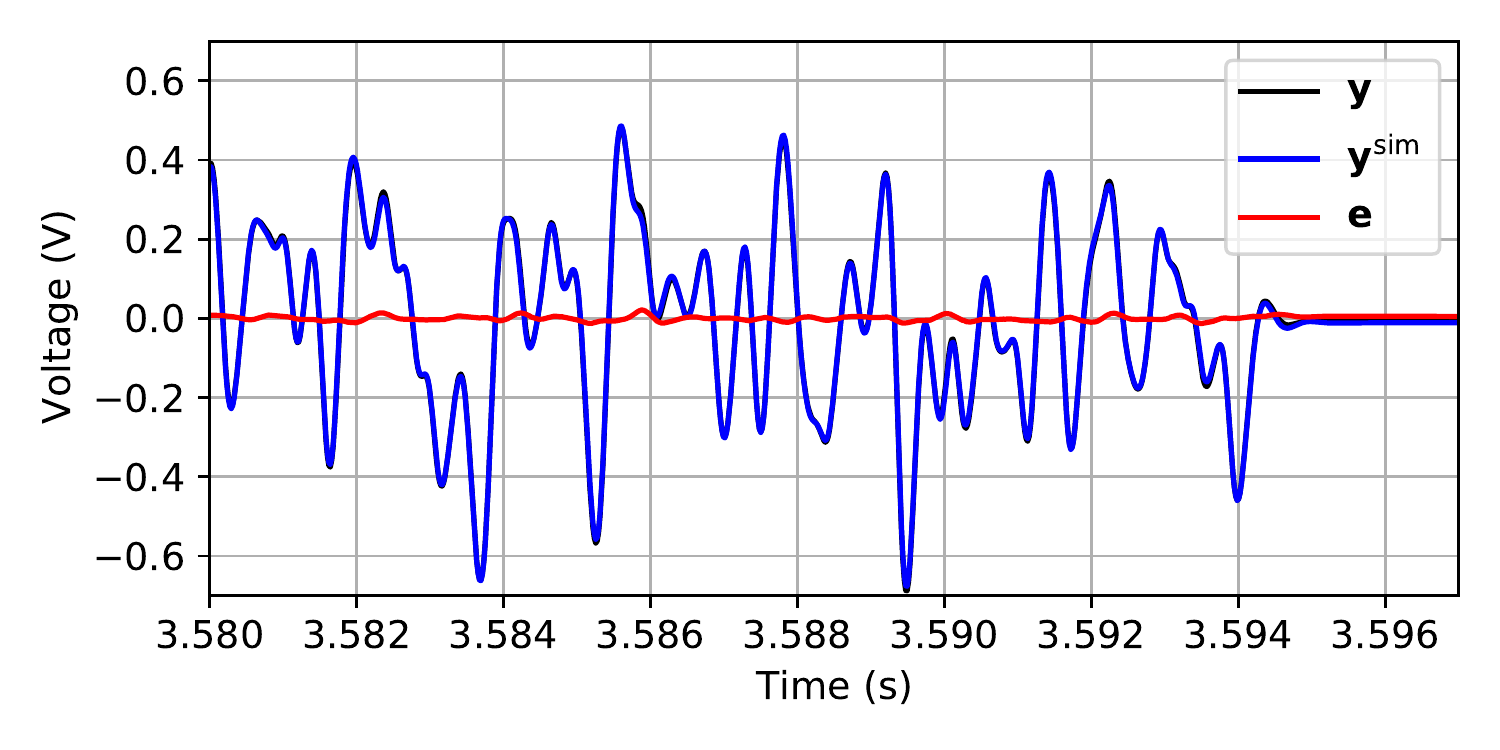} \vspace{-0.55cm}
	\caption{WH benchmark: measured output $\tvec{y}$ (black), simulated output $\tvec{y}^{\simul}$ (blue), and simulation error $\tvec{e}\! =\! \tvec{y}\! -\! \tvec{y}^{\simul}$ (red) on the  test dataset. 
}
	\label{fig:WH_timetrace}
\end{figure}
\begin{figure}
	\centering
	\includegraphics[width=.8\textwidth]{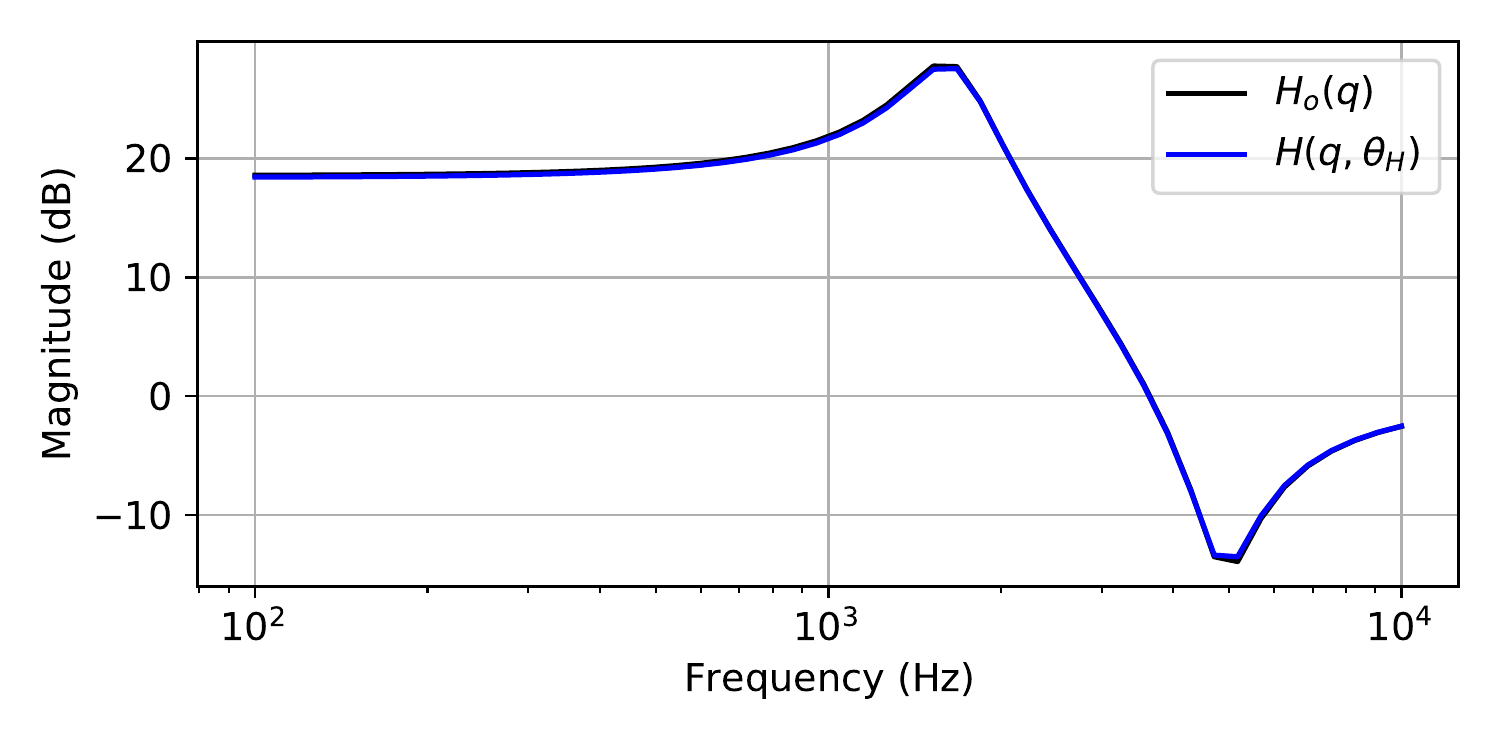} \vspace{-0.4cm}
	\caption{WH benchmark: Magnitude Bode plot of the filter $H_{\rm o}(q)$ and of the estimated filter $H(q,\theta_H)$.}
	\label{fig:WH_H}
\end{figure}

On the test dataset ($\nsamp=87000$ samples), the performance of the identified \Name \ model    are $\rm fit\!=\!96.9\%$ and $\rm{RMSE}\!=\!7.5$~mV. 
The measured output $\tvec{y}$ and simulated output $\tvec{y}^{\simul}$ on the test dataset are shown in Fig~\ref{fig:WH_timetrace}, together with the simulation error $\tvec{e} = \tvec{y}\! -\! \hat{\tvec{y}}$.\footnote{
For the sake of visualization,   a portion of the test data is shown.}
The magnitude Bode plot of the noise filter $H_{\rm o}(q)$ and of the identified $H(q,  \theta_H)$ are reported in Fig.~\ref{fig:WH_H}. The obtained results  show the capabilities of the \emph{deepPEM} in accurately reconstructing the system's output and the noise spectrum.

\subsection{PWH with quantized observations}
The experimental setup used in the second benchmark is an electronic circuit with a two-branch parallel Wiener-Hammerstein
structure described in \cite{schoukens2015parametric}. 
The original training dataset of the PWH benchmark  consists in 100 input/output training sequences, each one containing $N=16384$ samples. 
The 100 training sequences 
are obtained by exciting the system with random-phase multisine input signals at 5 different \emph{rms} levels: $\{100, 325, 550, 775, 1000\}$~mV, for 20 different realizations of the random phases. 
In this paper, we modify the original training dataset by discretizing the measured output voltage in 
$12$ equally spaced bins in the range $[-1 \ \  1]~V$. It is worth pointing out that in the benchmark datasets associated to \emph{rms} levels equal to $100$~mV and $325$~mV, the chosen quantization intervals lead to quantized outputs that only take $2$ different values. 

As a simulation model $\mathcal{M}(\cdot,\theta_\mathcal{M})$ (see Fig.~\ref{fig:quantized}) we use a sequential \Name\ network corresponding to the PWH structure and  constructed as the cascade connection of: a one-input-two-output $G$-block; two independent one-input-one-output feed-forward neural networks; and a two-input-one-output $G$-blocks. The $G$-blocks are characterized by $n_a=12$, $n_b=12$, and $n_k=1$.  

\begin{figure}
 \centering
 \includegraphics[width=.8\textwidth]{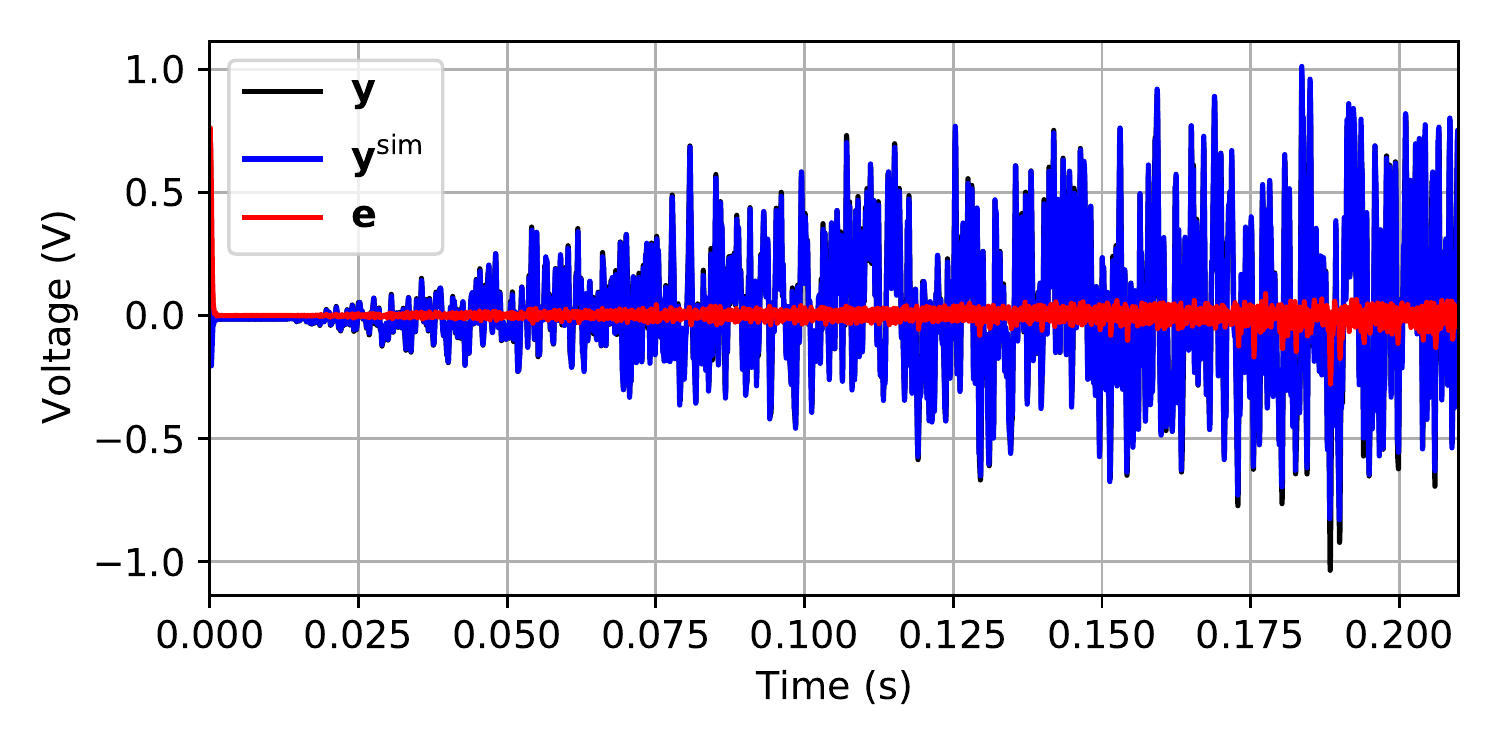} \vspace{-0.5cm}
 \caption{PWH benchmark: measured output $\tvec{y}$ (black),  simulated output $\tvec{y}^{\simul}$ (blue), and simulation error $\tvec{e}\! =\! \tvec{y}\! -\! \tvec{y}^{\simul}$ (red) on the growing envelope test dataset.}
 \label{fig:PWH_timetrace}
\end{figure}

We train the \Name\ network by maximizing the log-likelihood $\mathcal{L}(\theta)$ (Eq. \eqref{eq:likelihood_quant}) on the whole training dataset over $n=4000$ iterations with learning rate $\lambda=10^{-3}$. 
In the benchmark, 6 test datasets are provided. In the first 5 test datasets, the input signals are random-phase multisine at the same \emph{rms} levels considered in training (but with independent phase realizations), while in the latter the input signal is filtered Gaussian noise with an envelope that grows linearly over time.
On the 6 test datasets, the model achieves $\rm fit$ index $\{91.1,93.5,94.1,94.1,93.3,91.9\}\%$ and RMSE  
$\{3.65,8.20,12.02,16.04,22.06,21.37\}$~mV.  Time traces of the measured and simulated \Name\ output on the growing envelope test dataset are shown in Fig.~\ref{fig:PWH_timetrace}.

\section{Conclusions}
We have described  the operations required to make the linear time-invariant transfer functions fully compatible with the back-propagation algorithm.
This enables training of structured models combining transfer functions with other differentiable operators (such as deep neural networks) using standard deep learning software. Furthermore,  \emph{dynoNet}   allows us to train dynamical models without using back-propagation through time.

Furthermore, we have illustrated two  applications of the back-propagation-compatible linear dynamical block in system identification, namely the extension of the prediction error method to non-linear neural models in the presence of additive colored noise and the estimation of block-oriented models from quantized output observations.

Current  research is devoted to the analysis of neural model structures containing transfer functions through  linear system theory, to applications of these networks in other domains such as state estimation and time series classification, and to the extension of the $G$-block to the case of linear dynamics with parameter-varying coefficients.

\section*{Acknowledgments}
This work was partially supported by the European H2020-CS2 project ADMITTED, Grant agreement no. GA832003.

\bibliographystyle{plain}
\bibliography{ms}             
\end{document}